\documentclass[pdflatex,sn-mathphys,iicol]{sn-jnl}
\jyear{2021}%

\theoremstyle{thmstyleone}%
\theoremstyle{thmstyletwo}%

\theoremstyle{thmstylethree}%

\usepackage{graphicx}
\usepackage{subfigure}
\usepackage{amsmath}
\usepackage{amssymb}
\usepackage{dsfont}
\usepackage{bm}
\usepackage{booktabs}
\usepackage{wrapfig}
\usepackage{multirow}
\usepackage{url}

\usepackage{enumitem}
\def\etal{\emph{et al.}}
\usepackage{color}

\begin{document}

\title[Article Title]{NeRFTAP: Enhancing Transferability of Adversarial Patches on Face Recognition using Neural Radiance Fields}


\author[1,2]{\fnm{Xiaoliang} \sur{Liu}}\email{xiaoliang\_liu@smail.nju.edu.cn}

\author*[1,3]{\fnm{Furao} \sur{Shen}}\email{frshen@nju.edu.cn}
\author[1,2]{\fnm{Feng} \sur{Han}}\email{fenghan@smail.nju.edu.cn}
\author[1,4]{\fnm{Jian} \sur{Zhao}}\email{jianzhao@nju.edu.cn}
\author[1,2]{\fnm{Changhai} \sur{Nie}}\email{changhainie@nju.edu.cn}

\affil[1]{\orgdiv{State Key Laboratory for Novel Software Technology},  \orgname{Nanjing University}, \orgaddress{\country{China}}}

\affil[2]{\orgdiv{Department of Computer Science and Technology}, \orgname{Nanjing University}, \orgaddress{\country{China}}}
\affil[3]{\orgdiv{School of Artificial Intelligence}, \orgname{Nanjing University}, \orgaddress{\country{China}}}
\affil[4]{\orgdiv{School of Electronic Science and Engineering}, \orgname{Nanjing University}, \orgaddress{\country{China}}}


\abstract{Face recognition (FR) technology plays a crucial role in various applications, but its vulnerability to adversarial attacks poses significant security concerns. Existing research primarily focuses on transferability to different FR models, overlooking the direct transferability to victim's face images, which is a practical threat in real-world scenarios. In this study, we propose a novel adversarial attack method that considers both the transferability to the FR model and the victim's face image, called NeRFTAP. Leveraging NeRF-based 3D-GAN, we generate new view face images for the source and target subjects to enhance transferability of adversarial patches. We introduce a style consistency loss to ensure the visual similarity between the adversarial UV map and the target UV map under a 0-1 mask, enhancing the effectiveness and naturalness of the generated adversarial face images. Extensive experiments and evaluations on various FR models demonstrate the superiority of our approach over existing attack techniques. Our work provides valuable insights for enhancing the robustness of FR systems in practical adversarial settings.}
% \begin{keyword}
% %% keywords here, in the form: keyword \sep keyword
% Face recognition \sep adversarial attacks \sep transferability \sep NeRF \sep style consistency.
% \end{keyword}

\maketitle

%\IEEEraisesectionheading{\section{Introduction}\label{sec:introduction}}

\section{Introduction}
\label{sec:introduction}
Face recognition (FR) is a ubiquitous technology with a wide range of applications in our daily lives, including tasks such as phone unlocking, payment verification, and individual identification. The advent of deep learning has catalyzed significant advancements in FR~\cite{schroff2015facenet,wang2018cosface,deng2019arcface}. However, recent research indicates that deep neural network-based models are vulnerable to adversarial examples~\cite{goodfellow2014explaining,madry2018towards,dong2018boosting}—carefully crafted inputs that mislead the models and result in inaccurate predictions. This vulnerability poses a substantial threat to the security and reliability of FR models and systems. Hence, comprehensive research on adversarial attacks on FR is of paramount importance, offering invaluable insights to enhance the robustness of FR models and systems.

\begin{figure}
    \centering
    \includegraphics[width=1.0\linewidth]{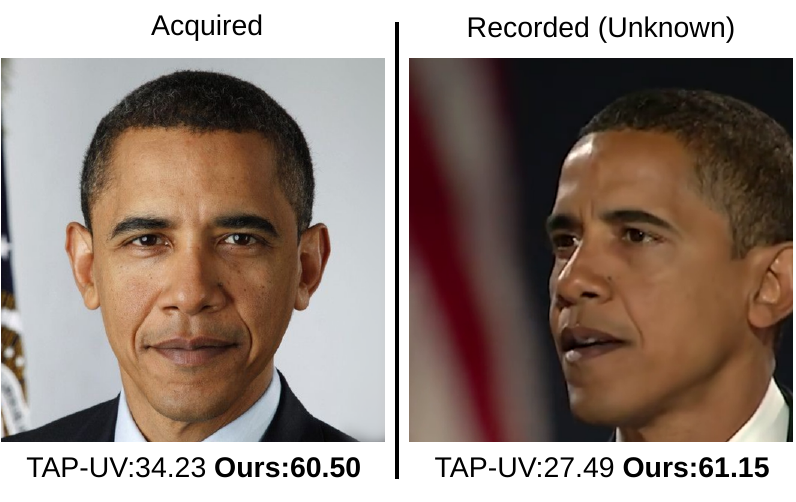}
    \caption{An example of an acquired and recorded face image of a victim. The confidence score comparison results between the TAP~\cite{xiao2021improving} method and our proposed NeRFTAP method are presented below each image. The confidence scores are obtained from the Tencent FR systems API~\cite{tencent}. The training target FR model utilizes ArcFace~\cite{deng2019arcface}. The confidence scores of the acquired target image of our method outperform TAP, indicating that our method generates more transferable adversarial patches for the FR model and the system. Moreover, the confidence scores of the recorded target images of our method also surpass TAP, demonstrating that our method exhibits greater transferability not only for the FR model and system but also for the victim's face images.}
    \label{fig:1}
\end{figure}

Extensive efforts have been dedicated to studying adversarial attacks on FR, categorizing them into evasion attacks and impersonation attacks, further divided into white-box and black-box attacks. Among these attacks, black-box-based impersonation poses the most severe and challenging threat to FR models and systems. While previous research~\cite{yang2021towards,xiao2021improving,liu2022rstam} has yielded promising results in this area, existing approaches have primarily focused on the transferability of adversarial examples to the FR model, disregarding their transferability to the victim's face image. In practical scenarios,  acquiring a face image of a victim is straightforward, but it does not necessarily correspond to the image recorded in the FR system, as illustrated in Fig.~\ref{fig:1}. The confidence score of the Transferable Adversarial Patch (TAP)~\cite{xiao2021improving} on the recorded face image, as shown in Fig.~\ref{fig:1}, decreases compared to the acquired image, indicating its inefficacy in transferring to the victim's face image.

To address this gap, our study considers both the transferability of adversarial examples to the FR model and the victim's face image. To address these limitations, we propose a novel black-box adversarial attack method called NeRFTAP. First, we utilize NeRF-based 3D-GAN to acquire NeRF hidden coding of source and target by inversion of 3D-GAN. Then, based on the Nerf hidden coding of the source and target, we can acquire the new perspective of the face of the source and target by simply inputting a new pose, which improves the transferability of the FR model and the victim's face image at the same time. We also introduce a style consistency loss to ensure the visual similarity between the adversarial UV map and the target UV map under a 0-1 mask. By jointly optimizing the adversarial UV map and the style consistency, we achieve robust impersonation attacks on FR systems.
In summary, the contributions of this work are as follows:
\begin{enumerate}
\item We propose NeRFTAP, a novel adversarial attack method on face recognition that considers transferability to both the FR model and victim's face images.
\item For the first time, we propose to utilize NeRF-based 3D-GAN to simultaneously improve the relocatability of adversarial patches on FR models and victim's face images.
\item We introduce a style consistency loss to ensure visually similar adversarial face images, making our attacks more effective and practical in real-world scenarios.
\item We demonstrate the efficacy of our approach through extensive experiments and evaluations on various FR models, highlighting its superiority over existing attack techniques.
\end{enumerate}

The remainder of this paper is organized as follows: In Section 2, we present the background and related works on adversarial attacks in FR. Section 3 details our proposed method, including the EG3D inversion process and the adversarial attack formulation. In Section 4, we provide experimental results and analysis. Finally, in Section 5, we conclude the paper and discuss potential future research directions.

% Leveraging recent developments in Neural Radiance Fields (NeRF)~\cite{mildenhall2021nerf} and Generative Adversarial Networks (GAN)~\cite{goodfellow2020generative}, we propose an EG3D-based adversarial patch attack method. A critical aspect of our method involves utilizing a single victim's face image to generate images from different viewpoints through EG3D inversion, thereby enhancing the transferability of adversarial patches to both the FR model and the victim's face image.

\section{Related Works}
In this section, we review some of the related works on adversarial attacks, adversarial attacks on face recognition, neural radiance fields and efficient geometry-aware 3D GAN.
\subsection{Adversarial Attacks}
Adversarial attacks represent a class of malicious techniques aimed at undermining the performance and reliability of machine learning models by introducing imperceptible perturbations to the input data. These attacks can be classified based on the attacker's knowledge of the target model, the attacker's goals, and the attacker's access to the input data, distinguishing between white-box or black-box attacks, targeted or untargeted attacks, and on-line or off-line attacks, respectively. In this subsection, we provide an overview of several prominent adversarial attack methods, including the Fast Gradient Sign Method (FGSM)~\cite{goodfellow2014explaining}, Projected Gradient Descent (PGD)~\cite{madry2018towards}, Momentum Iterative Method (MIM)~\cite{dong2018boosting}, Diverse Input Method (DIM)~\cite{xie2019improving}, and Translation-Invariant Method (TIM)~\cite{dong2019evading}.

The FGSM is a one-step gradient-based attack that crafts adversarial examples by introducing perturbations scaled by the sign of the loss function gradient with respect to the input. In contrast, the PGD extends the FGSM to an iterative process, projecting the perturbed input onto an $\ell_p$-norm ball centered at the original input. The MIM builds upon the PGD, incorporating momentum during the gradient calculation to improve robustness and escape from suboptimal solutions. To enhance diversity and transferability of adversarial examples, the DIM applies random transformations to the input before computing the gradient. The TIM exploits the translation-invariant property of convolutional neural networks to generate spatially robust adversarial examples.

\subsection{Adversarial Attacks on Face Recognition}
FR systems have become indispensable in various applications, rendering them susceptible to adversarial attacks. The primary objective of adversarial attacks on FR is to manipulate the decision-making process of the system, resulting in misclassifications or unauthorized access. These attacks can be broadly classified into two categories: evasion attacks and impersonation attacks.

Evasion attacks, often referred to as dodging attacks, attempt to reduce the similarity confidence between same-identity pairs, leading to misclassifications and a decrease in system accuracy. Conversely, impersonation attacks aim to enhance the feature similarity between the target individual and the attacker's source image of a different identity, enabling unauthorized access to the FR system. Due to the technical complexity involved, impersonation attacks pose a more challenging task than evasion attacks. As a result, our focus primarily revolves around studying impersonation attacks.

Numerous studies have been proposed to launch impersonation attacks on FR systems. Global-based methods, that is, noise-based methods like AdvFaces~\cite{deb2020advfaces}, employ a genetic algorithm to generate almost imperceptible perturbations covering the face globally. In contrast, LowKey~\cite{cherepanova2021lowkey} uses adversarial training to boost transferability. TIP-IM~\cite{yang2021towards} enhances the transferability of adversarial examples using multiple face models and random input transformations.  Patch-based methods such as Accessorize~\etal~\cite{sharif2016accessorize} and AdvHat~\cite{komkov2021advhat} innovatively use adversarial patches, like eyeglass frames or hats. In the physical world, these patches have demonstrated effectiveness in their attacks. TAP~\cite{xiao2021improving} and RSTAM~\cite{liu2022rstam} improve the transferability of adversarial patches by diversifying inputs. TAP achieves this by generating patches with various attributes and styles to adapt to different facial features and scenarios, while RSTAM uses random similarity transformations to adapt to different facial poses and illuminations. These studies highlight the vulnerabilities of FR systems and emphasize the importance of robust defensive mechanisms for securing practical applications of face recognition technology.

Nevertheless, while these studies have improved the transferability of adversarial examples, they have predominantly focused on FR models, largely overlooking the transferability to victim's face images.

\subsection{Neural Radiance Fields and Efficient Geometry-aware 3D GAN}
Neural Radiance Fields (NeRF)~\cite{mildenhall2021nerf}, a groundbreaking representation for 3D scenes, employs a continuous function parameterized by a neural network. This function maps 3D coordinates and viewing directions to color and density values. Despite its remarkable results in photorealistic view synthesis and 3D reconstruction, NeRF requires a vast number of input images and poses and is computationally demanding to train and query.

Efficient Geometry-aware 3D GAN (EG3D)~\cite{chan2022efficient} is a recent work that proposes an efficient geometry-aware 3D GAN framework that leverages NeRF as an inductive bias for 3D-aware image synthesis. EG3D uses a hybrid explicit-implicit tri-plane representation that stores features on axis-aligned planes that are aggregated by a lightweight implicit feature decoder for efficient volume rendering. EG3D also uses a pose-aware convolutional generator and a dual discriminator to achieve high-quality and view-consistent image synthesis and shape generation from 2D photographs without any explicit 3D or multi-view supervision.

\section{Method}
In this section, we provide an overview of the preliminary concepts and techniques that form the foundation of our proposed NeRFTAP method.
\subsection{Preliminaries}
In this study, we employ the EG3D framework proposed by Chan~\etal~\cite{chan2022efficient} to obtain the new view face images of the source and target. The EG3D pipeline, as depicted in Fig.~\ref{fig:1}, comprises several key modules, including a tri-plane generator based on StyleGAN2~\cite{Karras2019stylegan2}, a tri-plane representation module, a tri-plane decoder, a volume renderer, and a super-resolution module. The incorporation of tri-plane features is a fundamental aspect of the EG3D pipeline, allowing a 2D convolutional neural network (CNN) decoder to extract essential attributes such as color $\boldsymbol{c}$, density $\sigma$, and additional features $\boldsymbol{f}_c$, which are utilized for volume rendering tasks.

The color image $\boldsymbol{x}_c$ is obtained by performing volume rendering along the ray $\boldsymbol{r}$ and aggregating the values of $\boldsymbol{c}$ and $\sigma$. This aggregation process can be described by the equation: 
\begin{equation}
    \boldsymbol{x}_c(\boldsymbol{r})=\int_{t_n}^{t_f} T(t) \sigma(\boldsymbol{r}(t))\boldsymbol{c}(\boldsymbol{r}(t),\boldsymbol{d})dt
\end{equation}
where $T(t)=\exp(-\int_{t_n}^{t} \sigma(\boldsymbol{r}(s))ds)$ represents the accumulated transmittance. $t_n$ and $t_f$ indicate the near and far bounds of the spatial sampling locations, respectively. $\boldsymbol{d}$ represents the viewing direction.

Due to $\boldsymbol{x}_c$ being a low-resolution image, a super-resolution module is applied to obtain the final image $\boldsymbol{x}_{cf}$ as follows:
\begin{equation}
    \boldsymbol{x}_{cf}=SuperRes(\boldsymbol{x}_c,\boldsymbol{x}_{f}),
\end{equation}
where $\boldsymbol{x}_f$ is a feature image derived from the rendering process by tracing over the features $\boldsymbol{f}_c$. This feature image captures additional information that contributes to the overall appearance and characteristics of the final image.

\begin{figure*}
	\centering
		\includegraphics[width=\textwidth]{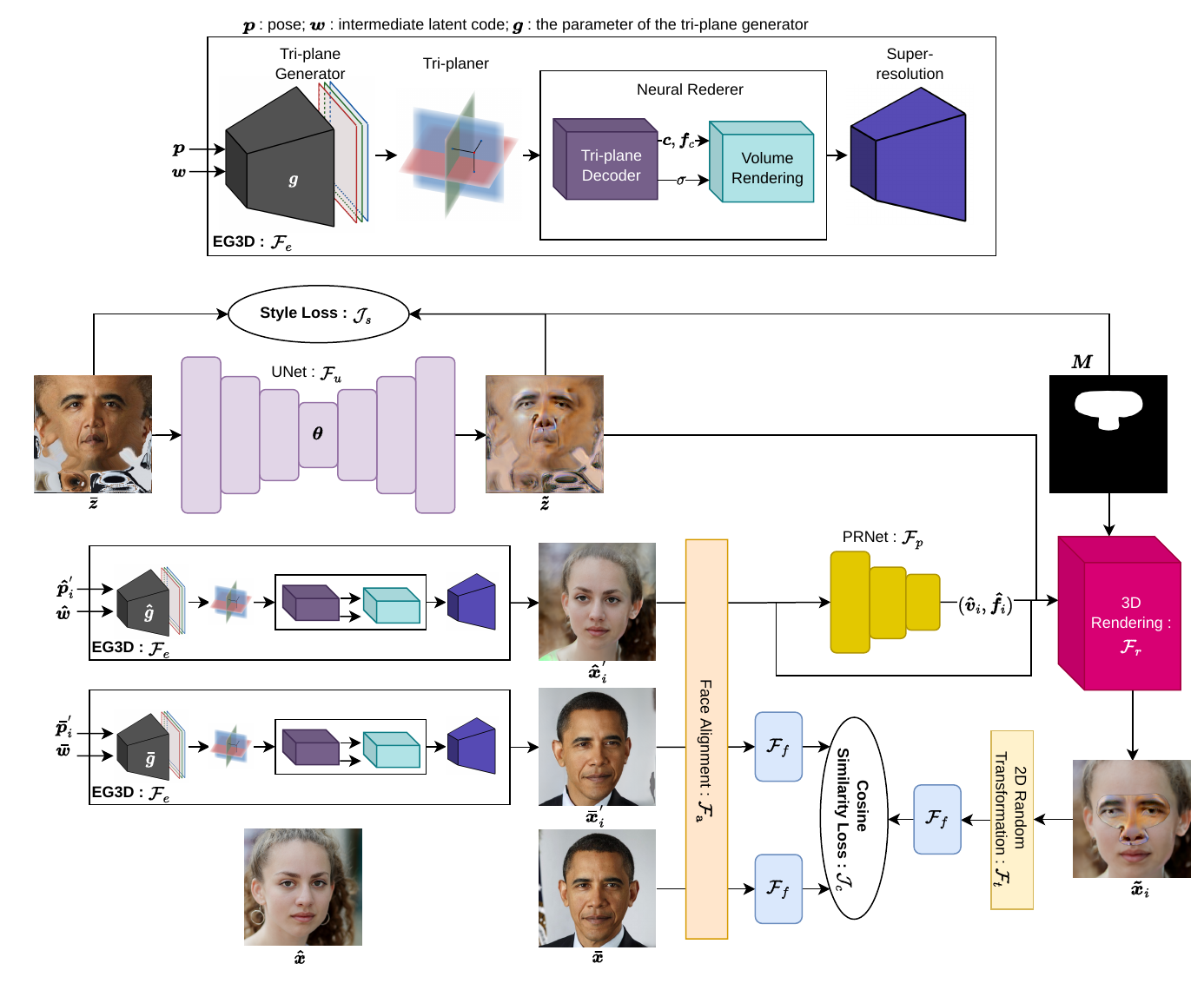}
	\caption{The overall framework of NeRFTAP. $\boldsymbol{\hat{x}}$ is the source face image. $\boldsymbol{\bar{x}}$ is the target face image. $\boldsymbol{\bar{z}}$ is the target UV map. $\mathcal{F}_f$ denotes the pre-trained FR model for extracting feature vectors. $\boldsymbol{\hat{p}}_i^{'}$ and $\boldsymbol{\bar{p}}_i^{'}$ are poses. $\boldsymbol{\hat{w}}$ and $\boldsymbol{\bar{w}}$ are  the intermediate latent codes of the target and source, respectively, obtained using EG3D inversion. $\boldsymbol{\hat{g}}$ and $\boldsymbol{\bar{g}}$ are the parameters of the tri-plane generator after fine-tuning using the source and target, respectively. $(\boldsymbol{\hat{v}}_{i},\boldsymbol{\hat{f}}_{i})$ are the mesh vertices and faces of $\mathcal{F}_a(\boldsymbol{\hat{x}}_i^{'})$. $\theta$ is the parameter of the UNet $\mathcal{F}_u$. $\boldsymbol{\tilde{z}}$ is the output adversarial UV map.  The objective is to obtain the adversarial UV map $\boldsymbol{\tilde{z}}$ by optimizing the parameter $\boldsymbol{\theta}$.}
	\label{fig:Overview}
\end{figure*}

\subsection{3D-GAN inversion}
Similar to Roich et al.'s PTI approach~\cite{roich2022pivotal}, our inversion process consists of two optimization steps: optimizing the intermediate latent code $\boldsymbol{w}$ and fine-tuning the parameters $\boldsymbol{g}$ of the tri-plane generator. Given an input image $\boldsymbol{x}$, we utilize DECA~\cite{DECA:Siggraph2021} to estimate its camera pose $\boldsymbol{p}$. The optimization objective for $\boldsymbol{w}$ can be defined as follows:
\begin{equation}
\begin{aligned}
\boldsymbol{w}_{x},\boldsymbol{n}=\mathop{argmin}_{\boldsymbol{w},\boldsymbol{n}}\mathcal{J}_{LPIPS}(\boldsymbol{x},\mathcal{F}_e(\boldsymbol{p},\boldsymbol{w},\boldsymbol{g})) \\ + \lambda_n \mathcal{J}_{n}(\boldsymbol{n}),
\end{aligned}
\end{equation}
where $\mathcal{F}_e$ represents EG3D. $\boldsymbol{g}$ denotes the pre-training parameters of the tri-plane generator. $\boldsymbol{n}$ is the noise vector as defined in~\cite{karras2020analyzing}. 
The perceptual loss $\mathcal{J}_{LPIPS}$ and the noise regularization term $\mathcal{J}_{n}$ with hyper-parameter $\lambda_n$ are used for optimization.

In the first optimization step, we freeze the generator, while in the subsequent step, we unfreeze the generator and perform fine-tuning with the objective:
\begin{equation}
\begin{aligned}
    \min_{\boldsymbol{g}} \mathcal{J}_{LPIPS}(\boldsymbol{x},\mathcal{F}_e(\boldsymbol{p},\boldsymbol{w}_x,\boldsymbol{g})) \\+\lambda_{L2} \mathcal{J}_{L2}(\boldsymbol{x},\mathcal{F}_e(\boldsymbol{p},\boldsymbol{w}_x,\boldsymbol{g})),
\end{aligned}
\end{equation}
where $\mathcal{J}_{L2}$ denotes the L2 loss, and $\lambda_{L2}$ is a hyper-parameter. During this fine-tuning stage, both the perceptual loss and L2 loss are considered to optimize the generator parameters $\boldsymbol{g}$.

\subsection{NeRFTAP}

Fig.~\ref{fig:Overview} illustrates the overall framework of NeRFTAP. Given the source face image $\boldsymbol{\hat{x}}$, the target face image $\boldsymbol{\bar{x}}$, and the 0-1 mask $\boldsymbol{M}$, we perform face alignment using $\mathcal{F}_a$, apply a 2D random transformation using $\mathcal{F}_t$, and perform 3D rendering using $\mathcal{F}_r$. The pre-trained FR model $\mathcal{F}_f$ extracts feature vectors, and PRNet\cite{feng2018joint} is denoted as $\mathcal{F}_p$, EG3D~\cite{chan2022efficient} as $\mathcal{F}_e$, and UNet~\cite{ronneberger2015u} as $\mathcal{F}_u$. The target UV map $\boldsymbol{\bar{z}}$ is obtained from $\boldsymbol{\bar{x}}$ using PRNet, and $\theta$ is the parameter of $\mathcal{F}_u$.

% We are given the source face image $\boldsymbol{\hat{x}}$, the target face image $\boldsymbol{\bar{x}}$ and the 0-1 mask $\boldsymbol{M}$. The operation $\mathcal{F}_a$ performs face alignment, $\mathcal{F}_t$ applies a 2D random transformation, and $\mathcal{F}_r$ performs 3D rendering. $\mathcal{F}_f$ denotes the pre-trained FR model for extracting feature vectors, $\mathcal{F}_p$ represents PRNet~\cite{feng2018joint}, $\mathcal{F}_e$ represents EG3D~\cite{chan2022efficient}, and $\mathcal{F}_u$ represents UNet~\cite{ronneberger2015u}. The target UV map $\boldsymbol{\bar{z}}$ is obtained from $\boldsymbol{\bar{x}}$ using PRNet. $\theta$ is the parameter of $\mathcal{F}_u$. 

Let 
\begin{equation}
\begin{aligned}
     &\boldsymbol{\tilde{z}} = \mathcal{F}_u(\boldsymbol{\bar{z}};\boldsymbol{\theta}),\\
     &\boldsymbol{\hat{x}}_i^{'}=\mathcal{F}_e(\boldsymbol{\hat{p}}_i^{'},\boldsymbol{\hat{w}},\boldsymbol{\hat{g}}),\\
     &\boldsymbol{\bar{x}}_{i}^{'}=\mathcal{F}_e(\boldsymbol{\bar{p}}_i^{'},\boldsymbol{\bar{w}},\boldsymbol{\bar{g}}),\\
     &\boldsymbol{\tilde{x}}_{i}=\mathcal{F}_r(\mathcal{F}_a (\boldsymbol{\hat{x}}_i^{'}),\underbrace{\mathcal{F}_{p}\mathcal{F}_a(\boldsymbol{\hat{x}}_i^{'})}_{(\boldsymbol{\hat{v}}_{i},\boldsymbol{\hat{f}}_{i})},\boldsymbol{\tilde{z}},\boldsymbol{M}),
\end{aligned}
\end{equation}
where $\boldsymbol{\hat{p}}_i^{'}$ and $\boldsymbol{\bar{p}}_i^{'}$ are poses. $\boldsymbol{\hat{w}}$ and $\boldsymbol{\bar{w}}$ are  the intermediate latent codes of the target and source, respectively, obtained using EG3D inversion. $\boldsymbol{\hat{g}}$ and $\boldsymbol{\bar{g}}$ are the parameters of the tri-plane generator after fine-tuning using the source and target, respectively. $(\boldsymbol{\hat{v}}_{i},\boldsymbol{\hat{f}}_{i})$ are the mesh vertices and faces of $\mathcal{F}_a(\boldsymbol{\hat{x}}_i^{'})$.  
$\mathcal{J}_c$ denotes the cosine similarity loss and $\mathcal{J}_s$ denotes the style loss. The objective is to obtain the adversarial UV map $\boldsymbol{\tilde{z}}$ by optimizing the parameter $\boldsymbol{\theta}$, and it is formulated as follows:
\begin{equation}
\small
\begin{aligned}
\min_{\boldsymbol{\theta}}\ \mathbb{E}_{\boldsymbol{\hat{p}}_i^{'} \sim \hat{\mathcal{S}}, \boldsymbol{\bar{p}}_i^{'} \sim \bar{\mathcal{S}}} [\mathcal{J}_{c}(\mathcal{F}_{f}\mathcal{F}_t(\boldsymbol{\tilde{x}}_{i}),\mathcal{F}_{f}\mathcal{F}_a(\boldsymbol{\bar{x}}_{i}^{'}),\mathcal{F}_{f}\mathcal{F}_{a}(\boldsymbol{\bar{x}}))\\ +\lambda_s \mathcal{J}_{s}(\boldsymbol{\bar{z}},\boldsymbol{\tilde{z}},\boldsymbol{M})]\\
\text{s.t.}\ \hat{\mathcal{S}} = \{\boldsymbol{\hat{p}} + \tau(2\boldsymbol{\beta}-1) \vert \boldsymbol{\beta} \sim \boldsymbol{Beta}(\alpha,\alpha) \} \\
 \bar{\mathcal{S}} = \{\boldsymbol{\bar{p}} + \tau(2\boldsymbol{\beta}-1) \vert \boldsymbol{\beta} \sim \boldsymbol{Beta}(\alpha,\alpha) \},
% &\begin{array}{r@{\ }r@{}l@{\ }l}
%      s.t. & \hat{\mathcal{S}} = \{\boldsymbol{\hat{p}} + \tau(2\boldsymbol{\beta}-1) | \boldsymbol{\beta} \sim \boldsymbol{Beta}(\alpha,\alpha) \}\\
%      & \bar{\mathcal{S}} = \{\boldsymbol{\bar{p}} + \tau(2\boldsymbol{\beta}-1) | \boldsymbol{\beta} \sim \boldsymbol{Beta}(\alpha,\alpha) \},
% \end{array}
\end{aligned}
\end{equation}
where $\lambda_s$ is a hyper-parameter. $\boldsymbol{\hat{p}}$ and $\boldsymbol{\bar{p}}$ are the poses of $\boldsymbol{\hat{x}}$ and $\boldsymbol{\bar{x}}$, respectively, generated using DECA~\cite{DECA:Siggraph2021}. $\boldsymbol{Beta}(\alpha, \alpha)$ represents a Beta distribution with hyper-parameter $\alpha$, and $\tau$ is a scaling factor. 

Let
\begin{equation}
\begin{aligned}
    \boldsymbol{\tilde{\nu}}_{i} = \mathcal{F}_{f}\mathcal{F}_t(\boldsymbol{\tilde{x}}_{i}), 
    \boldsymbol{\bar{\nu}}_{i}^{'} = \mathcal{F}_{f}\mathcal{F}_a(\boldsymbol{\bar{x}}_{i}^{'}),\boldsymbol{\bar{\nu}} =\mathcal{F}_{f}\mathcal{F}_{a}(\boldsymbol{\bar{x}}),
\end{aligned}
\end{equation}
and we define the cosine similarity loss $\mathcal{J}_c$ as follows: 
\begin{equation}
\mathcal{J}_c(\boldsymbol{\tilde{\nu}}_{i},\boldsymbol{\bar{\nu}}_{i}^{'},\boldsymbol{\bar{\nu}})=1-\frac{1}{2}(\frac{<\boldsymbol{\tilde{\nu}}_{i},\boldsymbol{\bar{\nu}}_{i}^{'}>}{\Vert\boldsymbol{\tilde{\nu}}_{i}\Vert \cdot \Vert\boldsymbol{\bar{\nu}}_{i}^{'} \Vert}+\frac{<\boldsymbol{\tilde{\nu}}_{i},\boldsymbol{\bar{\nu}}>}{\Vert\boldsymbol{\tilde{\nu}}_{i}\Vert \cdot \Vert\boldsymbol{\bar{\nu}}\Vert}),
\end{equation}
where $<\cdot,\cdot>$ denotes the inner product of the vectors.

We use feature maps extracted by a pretrained SqueezeNet~\cite{iandola2016squeezenet}, $\mathcal{F}_s$, to contruct the style loss following~\cite{gatys2016image}. The style loss $\mathcal{J}_s$ is defined as
\begin{equation}
\begin{aligned}
\mathcal{J}_s(\boldsymbol{\bar{z}},\boldsymbol{\tilde{z}},\boldsymbol{M})=\sum_{l=0}^{L} \gamma_l \sum_{j=0}^{C_l}\sum_{k=0}^{C_l}(\boldsymbol{A}_{l}[\mathcal{F}_s(\boldsymbol{\bar{z}}\odot \boldsymbol{M})]\\ -\boldsymbol{A}_{l}[\mathcal{F}_s(\boldsymbol{\tilde{z}}\odot \boldsymbol{M})])_{jk}^{2},
\end{aligned}
\end{equation}
where $L$ is the number of $\mathcal{F}_s$ layers and $C_l$ is the number of channels in activation. $\odot$ is the element-wise product. $\boldsymbol{A}_l[\cdot] \in \mathbb{R}^{C_l \times C_l}$ is the Gram matrix of features extracted at a set of $\mathcal{F}_s$ layers. $\gamma_l$ is the scalar weight to control the contribution of each layer when calculating the style loss. 

Additionally, the 2D random transformation uses a random similarity transformation, similar to RSTAM~\cite{liu2022rstam}, and the 3D rendering uses the PyTorch3D~\cite{ravi2020pytorch3d} renderer.

\section{Experiments}
In this section, we present the experimental results and analysis of our proposed NeRFTAP method. We evaluate its performance in comparison to other attack methods on various target FR models and systems, and we conduct an in-depth analysis of its effectiveness and transferability in impersonation attacks.
\subsection{Experimental Setup}
%\noindent\textbf{Datasets.}
\begin{figure*}
    \centering
    \includegraphics[width=0.9\linewidth]{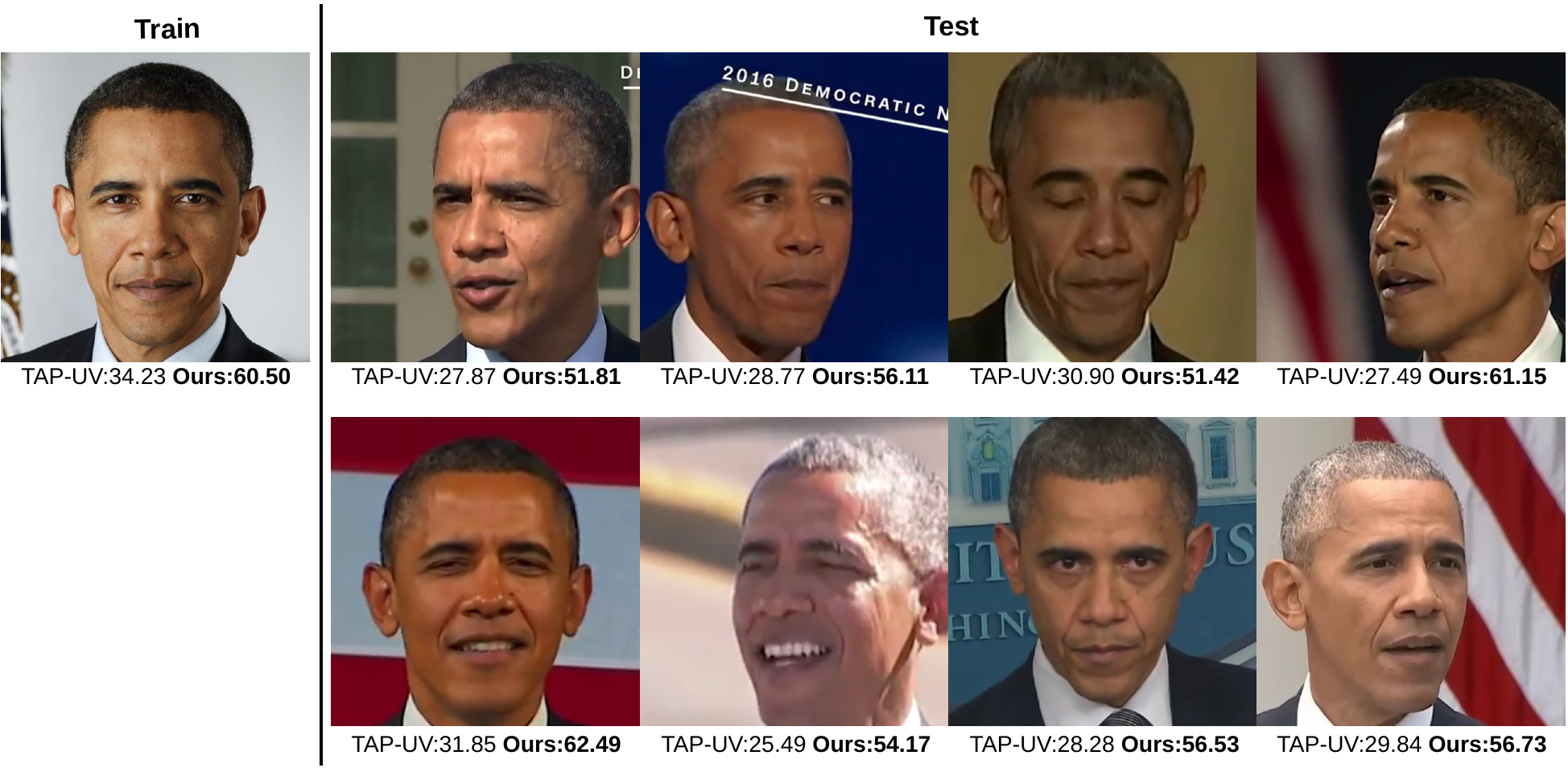}
    \caption{Examples of the victim's face images. The confidence score comparison results between the TAP~\cite{xiao2021improving} method and our proposed NeRFTAP method are presented below each image. The confidence scores are obtained from the Tencent FR systems API~\cite{tencent}. The training target FR model utilizes ArcFace~\cite{deng2019arcface}. }
    \label{fig:test}
\end{figure*}

\begin{figure*}
    \centering
    \subfigure[Eye]{
		\includegraphics[scale=0.3]{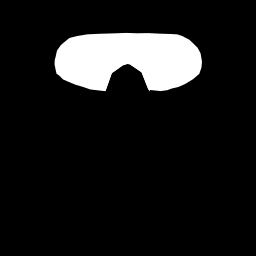}}
    \subfigure[Eye \& Nose]{
		\includegraphics[scale=0.3]{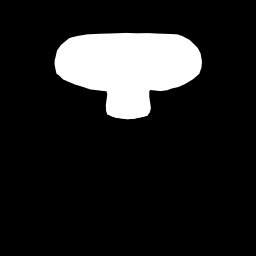}}
    \subfigure[Respirator]{
		\includegraphics[scale=0.3]{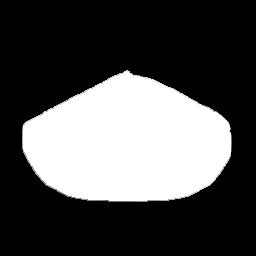}}
    \caption{Three 0-1 Masks, including Eye mask, Eye \& Nose mask, and Respirator mask.}
    \label{fig:0-1mask}
\end{figure*}

\textbf{Datasets.} The test set of face images for the targeted victim is collected from YouTube, comprising 8 diverse scenes, with each scene containing 20 face images displaying various poses, resulting in a total of 160 images. An illustrative example of these 8 scenes is shown in Fig.~\ref{fig:test}.  Additionally, we randomly select 10 source face images from the FFHQ~\cite{niemeyer2021campari} dataset, leading to a total of 1600 attack-target pairs for evaluation. In our experiments, we employ three different 0-1 masks, as shown in Fig.~\ref{fig:0-1mask}, including Eye mask, Eye \& Nose mask, and Respirator mask.

\textbf{Target FR Models and Systems.} For performance evaluation, we employ four different network architectures and renowned FR models, namely ArcFace~\cite{deng2019arcface}, CosFace~\cite{wang2018cosface}, MobleFace~\cite{chen2018mobilefacenets}, and ResNet50~\cite{he2016deep}. Additionally, we include two commercial FR systems, Face++~\cite{faceplusplus} and Tencent~\cite{tencent}, for further assessment.

\textbf{Baselines.}
As the baseline for comparison, we choose four global-based (noise-based) adversarial attack methods (FGSM~\cite{goodfellow2014explaining}, MIM~\cite{dong2018boosting}, DIM~\cite{xie2019improving}, and TIP-IM~\cite{yang2021towards}) and three patch-based adversarial attack methods (PGDAP~\cite{brown2017adversarial}, TAP~\cite{xiao2021improving}, and RSTAM~\cite{liu2022rstam}). To ensure a fair comparison, we extend the adversarial patches generated by the patch-based attack methods to the UV space using the method proposed by Alon~\etal~\cite{zolfi2022adversarial}.

\textbf{Evaluate Metrics.} 
%TODO:{deb2020advfaces,zhong2020towards,yin2021adv}
The success rate of impersonation attacks on FR models is evaluated using the Attack Success Rate (ASR)~\cite{deb2020advfaces,zhong2020towards,liu2022rstam}. ASR is computed as follows:
\begin{equation}
\begin{aligned}
&ASR=\\ &\frac{\sum_{i=1}^N \sum_{j=1}^M \mathbf{1}_{\epsilon} (cos[\mathcal{F}_f(\boldsymbol{\bar{x}}_i), \mathcal{F}_f({\boldsymbol{\tilde{x}}_j})] > \epsilon)}{N} \times 100\%,
\end{aligned}
\end{equation}
where $\boldsymbol{\bar{x}}$ refers to the target, and $\boldsymbol{\tilde{x}}$ represents the attack. $\mathcal{F}_f$ denotes the victim FR model. $N$ represents the number of targets and $M$ is the number of attacks. The function $\mathbf{1}_{\epsilon}$ denotes the indicator function, and $\epsilon$ is a predetermined threshold. The threshold $\epsilon$ for each victim FR model is determined by achieving a 0.001 False Acceptance Rate (FAR) on all possible image pairs in the LFW~\cite{huang2008labeled} dataset.

To evaluate attacks on FR systems, we use the mean confidence scores (MCS) as the evaluation metric. The MCS is calculated using the following formula:
\begin{equation}
MCS = \frac{\sum^N_{i=1} \sum_{j=1}^M \emph{conf}(\boldsymbol{\bar{x}}_i,\boldsymbol{\tilde{x}}_j)}{N} \times 100\%,
\end{equation}
where $\emph{conf}(\cdot,\cdot)$ represents the confidence score obtained from the FR system API for each pair of the target and the attack. The MCS provides an average measure of the confidence scores across the dataset, allowing for the assessment of the effectiveness of the attacks on the FR systems.

\textbf{Implementation Details.}
During the training process, we set the hyper-parameters $\lambda_s$, $\alpha$, and $\tau$ to be 0.0001, 0.2, and 15, respectively. Our training is conducted with a batch size of 8, and we run it for a total of 1000 epochs. For optimization, we initialize the learning rate to 0.0003 and utilize the Adam~\cite{kingma2014adam} optimizer with exponential decay rates set as $(\beta_1 ,\beta_2)=(0.9,0.999)$. Additionally, we employ a cosine annealing scheduler~\cite{loshchilov2016sgdr} technique for the learning rate scheduler.

\begin{table*}[]
\centering
\caption{Comparison results for our method and other attack methods on FR models and systems. ASR (\%) is used to evaluate impersonation attacks on FR models, while MCS (\%) is the metric for evaluating attacks on FR systems. $*$ denotes white-box attacks.}
\scalebox{1.0}{
\begin{tabular}{@{}c|cl|cccc|cc@{}}
\toprule
\multirow{2}{*}{\begin{tabular}[c]{@{}c@{}}Training \\ FR Model\end{tabular}} & \multicolumn{2}{c|}{\multirow{2}{*}{Method}} & \multicolumn{4}{c|}{Test FR Model} & \multicolumn{2}{c}{Test FR System} \\ \cmidrule(l){4-9} 
 & \multicolumn{2}{c|}{} & Arc. & Cos. & Mob. & Res. & Face++ & Tencent \\ \midrule
\multirow{7}{*}{ArcFace} & \multicolumn{1}{c|}{\multirow{4}{*}{Global}} & FGSM & 87.06* & 2.75 & 27.18 & 23.06  & 52.93 & 34.77 \\
 % & \multicolumn{1}{c|}{} & PGD & \textbf{100.00*} & 4.68 & 36.75 & 30.56 &  &  \\
 & \multicolumn{1}{c|}{} & MIM & \textbf{100.00*} & 4.00 & 35.62 &27.50  &53.29  & 28.92 \\
 & \multicolumn{1}{c|}{} & DIM & \textbf{100.00*} & 19.56 & 64.81 & 50.93 & 63.54 & 38.59 \\
 & \multicolumn{1}{c|}{} & TIP-IM & \textbf{100.00*} & 23.50 & 62.43 & 58.00 &64.48  & 40.48 \\ \cmidrule(l){2-9} 
 & \multicolumn{1}{c|}{\multirow{4}{*}{\begin{tabular}[c]{@{}c@{}}Eye \\ \&\\ Nose\end{tabular}}} & PGDAP-UV & 99.88* & 10.88 & 21.75 & 10.19  & 64.05 &32.90  \\
 & \multicolumn{1}{c|}{} & TAP-UV & 99.81* & 10.69 & 23.38 & 23.88  & 66.03 &31.41 \\
 & \multicolumn{1}{c|}{} & RSTAM-UV & 99.00* & 23.62 & 45.81 & 48.75 &  66.96 & 35.14 \\
 & \multicolumn{1}{c|}{} & Ours &  99.00*  &   \textbf{60.87} &  \textbf{70.25} &  \textbf{77.43} & \textbf{69.80} &  \textbf{55.97} \\ \midrule
\multirow{7}{*}{CosFace} & \multicolumn{1}{c|}{\multirow{4}{*}{Global}} & FGSM & 6.18 & 38.31* & 11.75 & 12.75 & 46.39 & 29.40 \\
 % & \multicolumn{1}{c|}{} & PGD & 11.56 & 99.00* & 10.00  & 5.06 &  &  \\
 & \multicolumn{1}{c|}{} & MIM & 12.75 & \textbf{99.06*} & 13.31  & 11.81 & 47.82 &26.98  \\
 & \multicolumn{1}{c|}{} & DIM & 38.18 & 98.00* & 50.12 & 30.68 & 57.47 & 34.89  \\
 & \multicolumn{1}{c|}{} & TIP-IM & 44.43 & 98.50* & 51.12 & 37.93 & 60.00 & 37.38  \\ \cmidrule(l){2-9} 
 & \multicolumn{1}{c|}{\multirow{4}{*}{\begin{tabular}[c]{@{}c@{}}Eye \\ \&\\ Nose\end{tabular}}} & PGDAP-UV & 5.56 & 98.75* & 17.56 & 13.81 & 63.70 & 35.96 \\
 & \multicolumn{1}{c|}{} & TAP-UV & 13.38 & 98.38* & 35.88 & 31.88 & 63.83 & 37.49 \\
 & \multicolumn{1}{c|}{} & RSTAM-UV & 43.12 & 93.00* & 64.19 & 64.75  & 66.15 &47.19  \\
 & \multicolumn{1}{c|}{} & Ours & \textbf{90.50} & 94.06* & \textbf{86.50} & \textbf{77.94} & \textbf{68.36} &  \textbf{54.08} \\ \midrule
\multirow{7}{*}{MobileFace} & \multicolumn{1}{c|}{\multirow{4}{*}{Global}} & FGSM & 23.93 & 2.62 & 92.81* & 38.37 & 51.85 & 33.38 \\
 % & \multicolumn{1}{c|}{} & PGD & 29.56 & 4.37 &  99.62* & 51.12 &  &  \\
 & \multicolumn{1}{c|}{} & MIM & 33.00 & 5.00 & \textbf{99.87*} & 47.37 & 49.90 & 28.12  \\
 & \multicolumn{1}{c|}{} & DIM & 56.18 & 35.31 & 99.43* & 76.25 & 56.37 & 37.67 \\
 & \multicolumn{1}{c|}{} & TIP-IM &50.93  &34.75  &99.37*  & 76.18 & 59.31 & 38.18 \\ \cmidrule(l){2-9} 
 & \multicolumn{1}{c|}{\multirow{4}{*}{\begin{tabular}[c]{@{}c@{}}Eye \\ \&\\ Nose\end{tabular}}} & PGDAP-UV & 32.19 & 29.62 & 99.38* & 44.75 & 66.97 & 41.22 \\
 & \multicolumn{1}{c|}{} & TAP-UV &69.62 & 43.31 & 98.50* & 72.94 & 70.44 & 45.22 \\
 & \multicolumn{1}{c|}{} & RSTAM-UV & 79.12 & 57.38 & 96.69* & 82.12 & 72.50 & 48.97 \\
 & \multicolumn{1}{c|}{} & Ours & \textbf{89.62} & \textbf{65.06} & 98.38* & \textbf{91.31}  & \textbf{75.96} &\textbf{58.09}  \\ 
 % \midrule
% \multirow{9}{*}{ResNet50} & \multicolumn{1}{c|}{\multirow{5}{*}{Global}} & FGSM & 4.93 & 2.31 & 18.68 & 62.81* &  &  \\
%  & \multicolumn{1}{c|}{} & PGD & 28.68 & 7.18 &65.00  & \textbf{100.00*}&  &  \\
%  & \multicolumn{1}{c|}{} & MIM & 31.12 & 9.75 & 71.31 & \textbf{100.00*} &  &  \\
%  & \multicolumn{1}{c|}{} & DIM & 50.93 &29.93  &76.87 & 99.81* &  &  \\
%  & \multicolumn{1}{c|}{} & TIP-IM & 44.00 &38.18  & 80.87 & 99.68* &  &  \\ \cmidrule(l){2-9} 
%  & \multicolumn{1}{c|}{\multirow{4}{*}{\begin{tabular}[c]{@{}c@{}}Eye \\ \&\\ Nose\end{tabular}}} & PGDAP-UV &21.87  &  21.00 & 43.62 & 99.75* &  &  \\
%  & \multicolumn{1}{c|}{} & TAP-UV & 47.56 & 35.87 &61.37  & 99.31* &  &  \\
%  & \multicolumn{1}{c|}{} & RSTAM-UV &64.00  &45.43  & 74.37 & 97.37* &  &  \\
%  & \multicolumn{1}{c|}{} & Ours & \textbf{83.93} & \textbf{63.31} & \textbf{89.62} & 99.75* &  &  \\  
 \bottomrule
\end{tabular}
}
\label{tab:1}
\end{table*}

\begin{figure*}
    \centering
    \subfigure[Face++]{
		\includegraphics[width=0.45\textwidth]{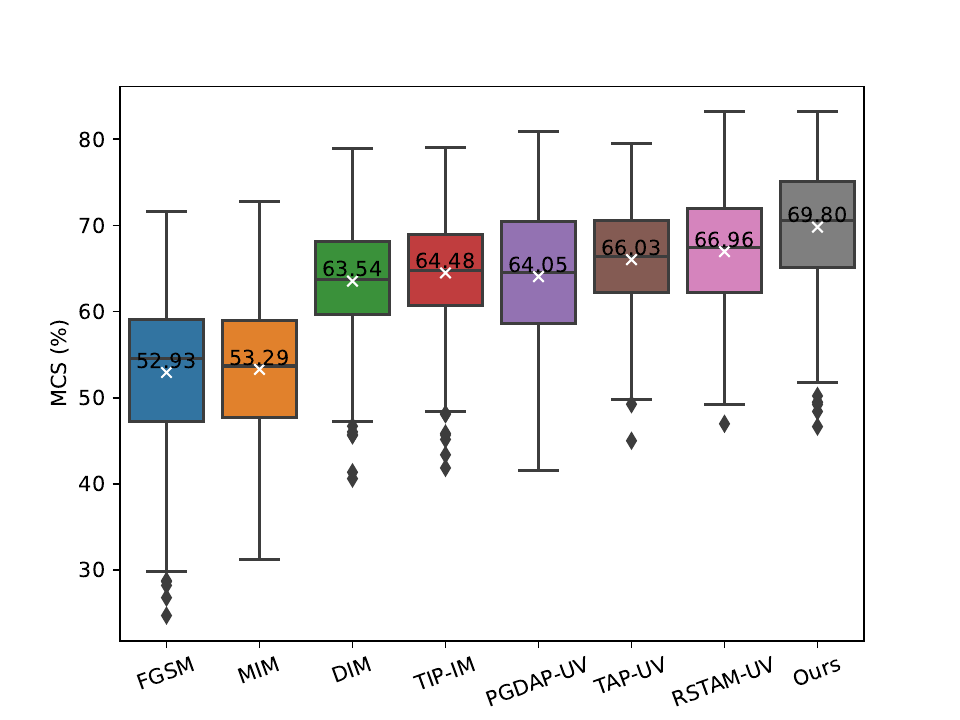}}
    \subfigure[Tencent]{
		\includegraphics[width=0.45\textwidth]{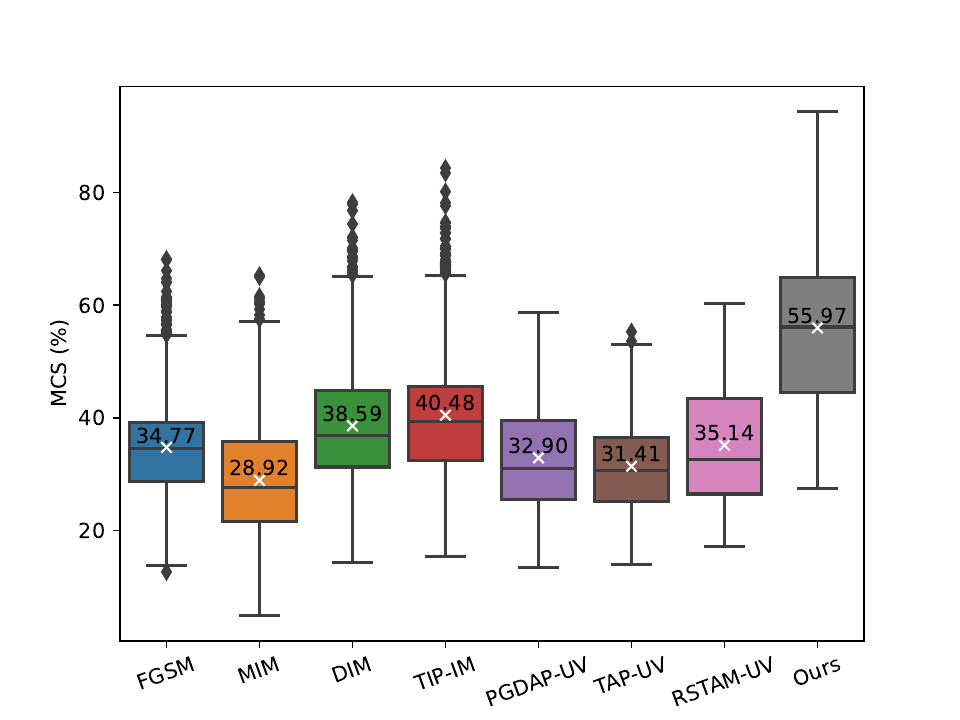}}
    \caption{Confidence scores returned from Face++ and Tencent. The training target FR model uses ArcFace.}
    \label{fig:mcs}
\end{figure*}

\begin{table*}[]
\centering
\caption{Comparison results on different 0-1 masks. $*$ indicates white-box attacks.}
\scalebox{0.95}{
\begin{tabular}{@{}c|l|cccc|cccc@{}}
\toprule
\multirow{2}{*}{\begin{tabular}[c]{@{}c@{}}Training \\ FR Model\end{tabular}} & \multicolumn{1}{c|}{\multirow{2}{*}{Method}} & \multicolumn{4}{c|}{Eye} & \multicolumn{4}{c}{Respirator} \\ \cmidrule(l){3-10} 
 & \multicolumn{1}{c|}{} & Arc. & Cos. & Mob. & Res.  & Arc. & Cos. & Mob. & Res. \\ \midrule
\multirow{4}{*}{ArcFace} & PGDAP-UV & \textbf{98.88*} & 2.31 & 11.50 & 5.88 & \textbf{97.44*} & 4.75 & 13.69 & 20.06 \\
 & TAP-UV & 98.06* & 1.56 & 10.62 & 6.38   & 96.31* & 5.69 & 13.00 & 30.12 \\
 & RSTAM-UV & 89.31* & 5.06 & 14.75 & 12.56 &  88.25* & 7.44 & 33.44 &  45.88\\
 & Ours & 91.50* & \textbf{20.25} & \textbf{25.31} & \textbf{31.44}  & 85.56* & \textbf{9.31} & \textbf{48.44} & \textbf{53.00} \\ \midrule
\multirow{4}{*}{CosFace} & PGDAP-UV & 2.12 & \textbf{97.50*} & 2.00 & 2.94 &   5.25 & \textbf{97.25*} & 11.00 & 12.50 \\
 & TAP-UV & 3.38 & 96.44* & 10.25 & 9.06 & 6.81 & 96.31* & 13.50 & 10.94\\
 & RSTAM-UV & 7.44 & 81.44* & 17.19 & 16.25 &  10.88 & 79.50* & 19.38 & 14.31\\
 & Ours & \textbf{ 33.75} & 83.94* & \textbf{24.56} & \textbf{26.56} &  \textbf{35.75} & 76.12* & \textbf{42.06} & \textbf{43.81} \\ \midrule
\multirow{4}{*}{MobileFace} & PGDAP-UV & 7.19 & 3.75 & \textbf{94.12*} & 21.31   & 21.62 & 10.88 & \textbf{95.38*} & 45.69\\
 & TAP-UV & 18.12 & 10.31 & 92.50* & 33.00 &   26.94 & 19.75 & 93.94* & 61.94\\
 & RSTAM-UV & 23.56 & 17.75 & 84.06* & 41.56 &35.94 & 22.75 & 88.75* & 70.12\\
 & Ours & \textbf{46.06} & \textbf{31.88} & 84.56* &  \textbf{54.06}& \textbf{51.25} & \textbf{30.06} & 90.56* & \textbf{75.31} \\
%  \midrule
% \multirow{4}{*}{ResNet50} & PGDAP-UV & 3.93 & 1.81 & 10.56 & \textbf{88.50*}   & 13.81 & 9.93 & 31.81 & \textbf{96.31*} \\
%  & TAP-UV & 8.87 & 6.43 & 19.18 & 85.87*   & 23.00 & 13.62 & 46.50 & 95.43* \\
%  & RSTAM-UV & 14.93 & 9.31 & 18.68 & 74.68*   & 30.62 & 14.56 &53.31  & 91.43* \\
%  & Ours & \textbf{33.75} & \textbf{24.62} & \textbf{46.06} & 85.25*  & \textbf{41.50} & \textbf{19.06} & \textbf{62.87} & 91.43* \\
 \bottomrule
\end{tabular}
}
\label{tab:2}
\end{table*}

\subsection{Comparison Study}

\textbf{Comparison with Other Attack Methods.} Tab.~\ref{tab:1} presents a comprehensive comparison between various attack methods, including our proposed NeRFTAP, and other existing approaches on different target FR models and systems. The perturbation boundaries for the global-based methods are uniformly set to 0.031. The results clearly demonstrate the consistent superiority of NeRFTAP over other methods in most scenarios, achieving higher ASR and MCS values.

In particular, when compared to the well-known TAP method, NeRFTAP exhibits exceptional performance in black-box attacks on all tested FR models and systems, showcasing its efficacy in impersonation attacks and remarkable transferability. These findings underscore NeRFTAP's ability to effectively manipulate FR systems and gain unauthorized access, establishing it as a potent adversarial attack method.

Additionally, the box plot analysis depicted in Fig.~\ref{fig:mcs} further reinforces our NeRFTAP method's advantage by consistently achieving higher confidence scores on both the Face++ and Tencent FR systems. Specifically, when evaluating the attack performance on the Tencent FR system, NeRFTAP's MCS outperforms TAP by 24.56\%, RSTAM by 20.83\%, and even the global-based method TIP-IM by 15.49\%. These remarkable results highlight NeRFTAP's exceptional effectiveness in attacking the Tencent FR system, surpassing alternative methods with significantly higher confidence scores. The higher confidence scores directly signify NeRFTAP's capability to adeptly manipulate the decision-making process of FR systems, leading to more successful and precise impersonation attacks. The robust performance demonstrated across two distinct FR systems further reaffirms the reliability and transferability of our NeRFTAP approach.

\begin{figure*}
    \centering
    \includegraphics[width=0.75\textwidth]{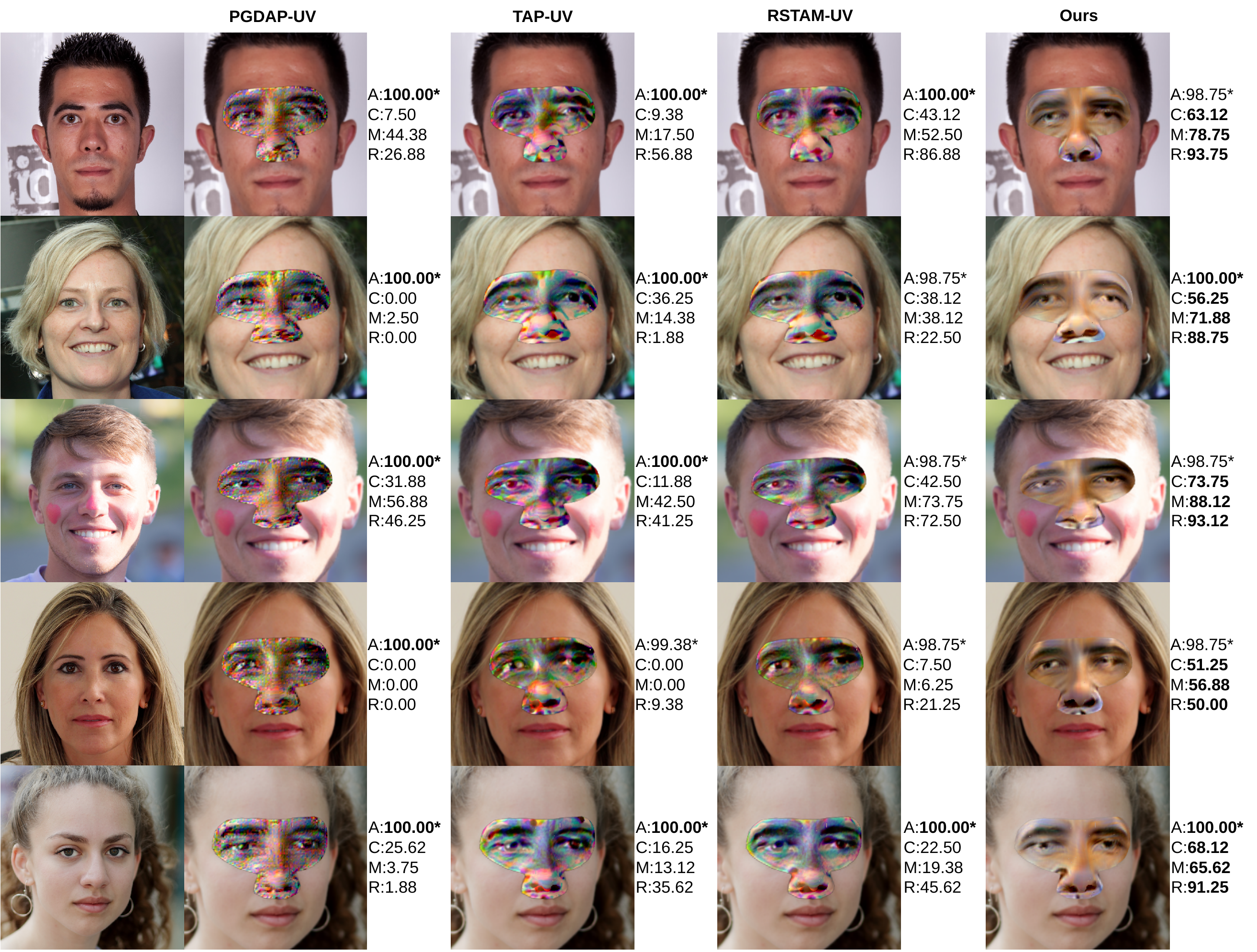}
    \caption{Comparative results of visual. The right of each image are the corresponding ASRs for different FR models (A: ArcFace, C: CosFace, M: MobileFace, R: ResNet50). $*$~indicates white-box attacks.}
    \label{fig:vis}
\end{figure*}

\begin{figure*}
    \centering
    \includegraphics[width=0.75\textwidth]{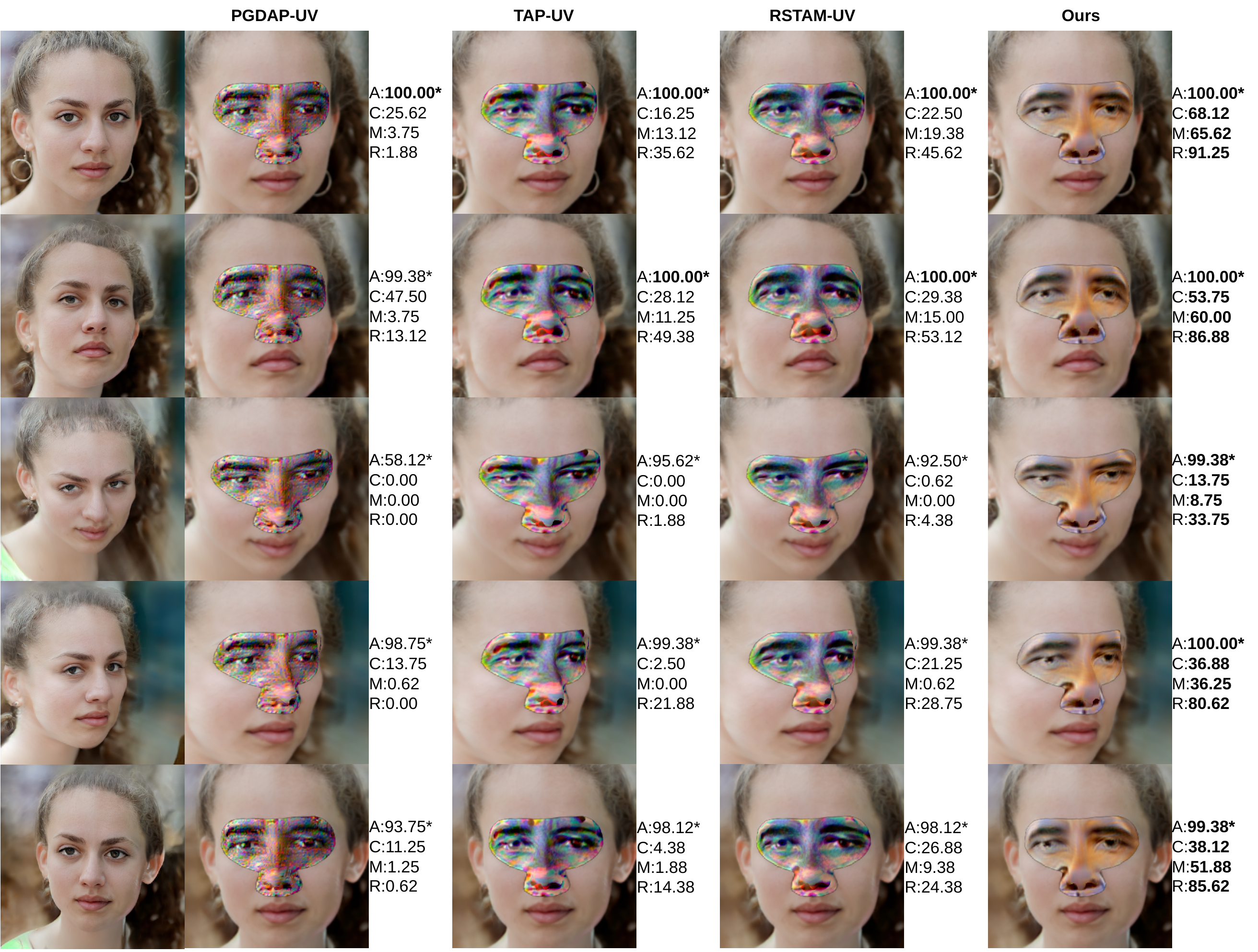}
    \caption{Comparison results for the source in different views. The right of each image are the corresponding ASRs for different FR models (A: ArcFace, C: CosFace, M: MobileFace, R: ResNet50). $*$ indicates white-box attacks.}
    \label{fig:newView}
\end{figure*}

\textbf{Comparison Results on Different 0-1 Masks.} Tab.~\ref{tab:2} showcases the comparison results for our NeRFTAP method and other attack techniques under the Eye and Respirator mask settings. Notably, in the black-box scenario, our methods consistently exhibit superiority over the alternative approaches. The comparison results in Tab.~\ref{tab:2} demonstrate that our NeRFTAP method consistently outperforms other attack techniques under both Eye and Respirator mask settings in the black-box scenario. This superior performance showcases the excellent transferability and effectiveness of our approach in manipulating face recognition models, regardless of the presence of different 0-1 masks.

\textbf{Comparative Results of Visual.}
Fig.~\ref{fig:vis} displays the comparative visual results of our NeRFTAP method, showcasing the adversarial examples generated for different FR models. The training target FR model utilizes ArcFace. The ASRs are provided to the right of each image, indicating the percentage of successful impersonation attacks on the respective FR models. As can be seen in Fig.~\ref{fig:vis}, our method is significantly superior to the other methods under the black-box attack setting, which also indicates that the transferability of the adversarial patches generated by our method is better.

Furthermore, Fig.~\ref{fig:newView} presents the comparison results for the source in different views, where our NeRFTAP method demonstrates its ability to maintain high ASRs across various FR models when the input face image is captured from diverse viewpoints. The ASRs for each scenario are shown beside the corresponding images, further illustrating the effectiveness and robustness of our approach against variations in facial pose and appearance.

\begin{table*}[]
\centering
\caption{Results of the ablation study. $*$ indicates white-box attacks.  }
\begin{tabular}{@{}l|llll@{}}
\toprule
          & Arc. & Cos. & Mob. & Res. \\ \midrule
Full      &    99.00*  &   \textbf{60.87}   &   \textbf{70.25}   &  \textbf{77.43}    \\
w/o Style Loss &   99.43*   &  43.25    &  49.00    &  52.37    \\
w/o 2D Random Tran. &    \textbf{99.68*}  &   39.37   &  58.00    &  65.62    \\
w/o EG3D &   99.31*   &   45.62   &    54.12  &  65.75    \\
 \bottomrule
\end{tabular}
\label{tab:3}
\end{table*}

\begin{figure*}
    \centering
    \subfigure[Hyper-parameter $\alpha$]{
		\includegraphics[width=0.4\textwidth]{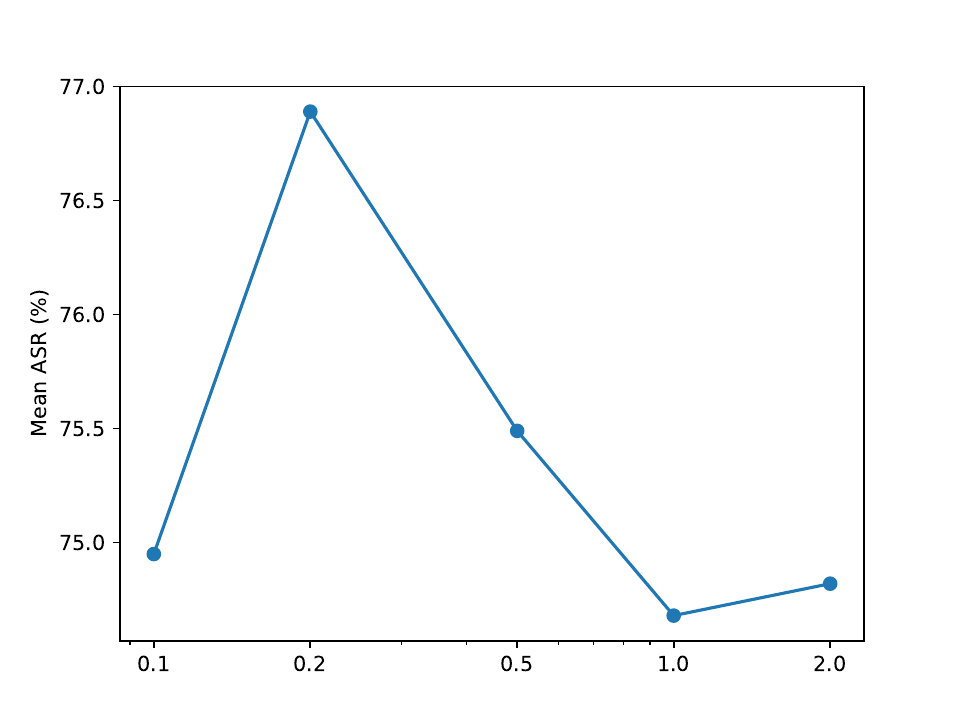}}
    \subfigure[Scaling Factor $\tau\ (^\circ)$]{
		\includegraphics[width=0.4\textwidth]{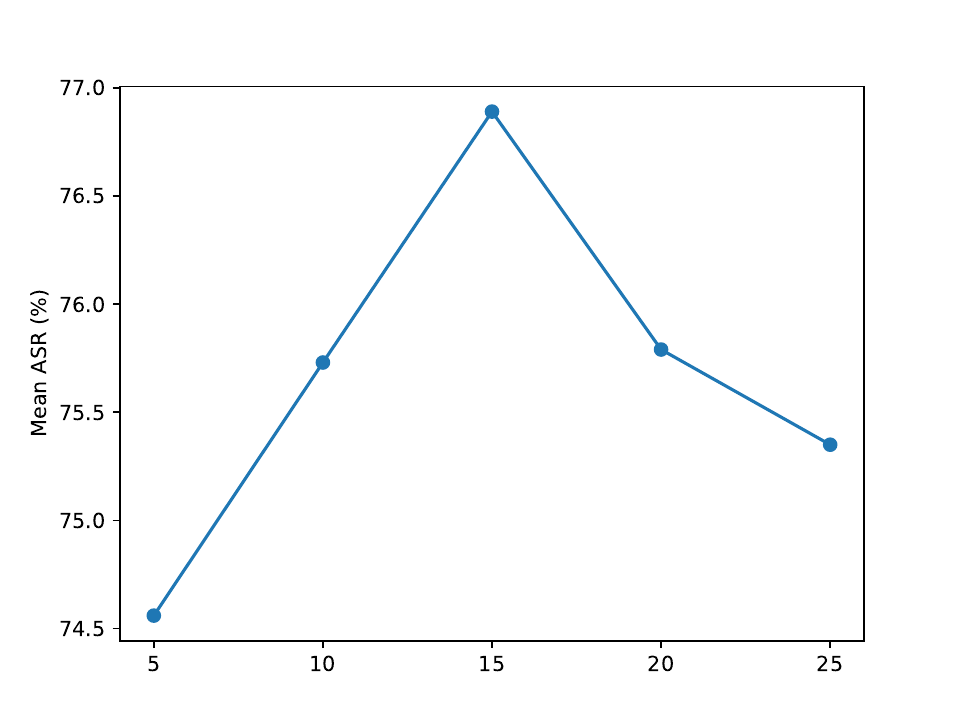}}
    \caption{Results of hyper-parameters analysis. }
    \label{fig:hp}
\end{figure*}

\subsection{Ablation Study}
Tab.~\ref{tab:3} presents the results of the ablation study, where we evaluate the impact of different components in our NeRFTAP method. The "Full" configuration represents our complete NeRFTAP method, which achieved high attack success rates on all four tested FR models. By comparing with the ``Full" configuration, we observe that removing the "Style Loss" leads to a decrease in attack success rates, indicating that the style consistency constraint plays a crucial role in enhancing the transferability of the adversarial patches. Similarly, when excluding the ``2D Random Transformation", the attack success rates drop notably, further demonstrating the significance of this augmentation in improving the patch's effectiveness in the black-box scenario.

Moreover, omitting the ``EG3D" component also results in decreased attack success rates, underscoring the importance of the enhanced gradient computation in generating more potent adversarial patches. The ablation study reaffirms that each component in our NeRFTAP method contributes synergistically to its outstanding performance in black-box attacks, making it a comprehensive and effective adversarial attack methodology in the realm of face recognition systems.

\subsection{Sensitivity Analysis of Hyper-parameters}
In this section, we conduct a sensitivity analysis of the hyper-parameters used in our NeRFTAP method. Specifically, we investigate the impact of the hyper-parameter $\alpha$ associated with the Beta distribution and the scaling factor $\tau$ on the performance of our adversarial attack approach.

As shown in Fig.~\ref{fig:hp}, we plot the results of the hyper-parameter analysis. The hyper-parameter $\alpha$ is examined in the range of values, and the scaling factor $\tau$ is varied accordingly. Among the different settings, we find that when $\alpha=0.2$ and $\tau=15^\circ$, our NeRFTAP method achieves the most favorable outcomes. These specific hyper-parameter values maximize the generation of realistic and transferable adversarial patches, leading to higher attack success rates on the target FR models.

The sensitivity analysis of hyper-parameters provides insights into the influence of these parameters on the effectiveness of our NeRFTAP method. Fine-tuning these hyper-parameters based on the specific application scenario can further enhance the performance and robustness of the attack methodology.

\section{Conclusions}
In this study, we introduced NeRFTAP, a novel black-box adversarial attack method that considers both the transferability to the FR model and the victim's face image. By leveraging NeRF-based 3D-GAN, we enhanced the transferability of adversarial patches by generating new view face images for the source and target subjects. Additionally, we incorporated a style consistency loss to ensure the visual similarity between the adversarial UV map and the target UV map under a 0-1 mask, resulting in more effective and natural adversarial face images.

Through extensive experiments and evaluations on various FR models, we demonstrated the superiority of NeRFTAP over existing attack techniques. Our method achieved robust impersonation attacks on FR systems, highlighting the practical threat of direct transferability to victim's face images in real-world scenarios.

In conclusion, our work contributes valuable insights to enhance the robustness of FR systems against adversarial attacks and provides a step towards strengthening the security of face recognition technology in face of emerging threats. Future research directions may include exploring other potential vulnerabilities and further improving the effectiveness and generalizability of adversarial attack methods.

\section*{Acknowledgments}
This work was supported in part by the STI 2030-Major Projects of China under Grant 2021ZD0201300, and by the National Science Foundation of China under Grant 62276127.

\section*{Declarations}
\noindent\textbf{Data Availability.}
The datasets generated and analysed during this study are available in the https://github.com/NVlabs/ffhq-dataset

\noindent\textbf{Conflict of Interest.} The authors declare that they have no conflict of interest.

\bibliography{references}

\vspace{1cm}
\end{document}